\documentclass[format=acmsmall, review=false, screen=false]{acmart}

\usepackage{booktabs} 
\pdfoutput=1
\usepackage[ruled]{algorithm2e} 

\SetAlFnt{\small}
\SetAlCapFnt{\small}
\SetAlCapNameFnt{\small}
\SetAlCapHSkip{0pt}
\IncMargin{-\parindent}

\usepackage{listings}
\begin{document}
\title[AITuning]{AITuning: Machine Learning-based Tuning Tool for Run-Time Communication Libraries}

\author{Alessandro Fanfarillo}
\orcid{0000-0003-3487-7452}
\affiliation{%
  \institution{National Center for Atmospheric Research}
  \city{Boulder}
  \country{Boulder, Colorado, USA}}

\email{elfanfa@ucar.edu}

\author{Davide Del Vento}
\affiliation{%
  \institution{National Center for Atmospheric Research}
  \city{Boulder}
  \country{Boulder, Colorado, USA}}
\email{ddvento@ucar.edu}

\begin{abstract}
  In this work, we address the problem of tuning communication libraries by using a deep reinforcement learning approach.
  Reinforcement learning is a machine learning technique incredibly effective in solving game-like situations.
  In fact, tuning a set of parameters in a communication library in order to get better performance in a parallel application
  can be expressed as a game: \textit{Find the right combination/path that provides the best reward}.
  Even though AITuning has been designed to be utilized with different run-time libraries,
  we focused this work on applying it to the OpenCoarrays run-time communication library, built on top of MPI-3.
  This work not only shows the potential of using a reinforcement learning algorithm for tuning communication libraries, but also
  demonstrates how the MPI Tool Information Interface, introduced by the MPI-3 standard, can be used effectively by run-time libraries
  to improve the performance without human intervention.
\end{abstract}

%
%


%
%

\keywords{MPI, Machine Learning, Reinforcement Learning, Coarray, Fortran}

\maketitle

\renewcommand{\shortauthors}{A. Fanfarillo et al.}

\section{Introduction}

Tuning a general-purpose communication library is tightly related to the communication pattern utilized
by the application, the network interconnect, the computer architecture, and the problem size.
Profilers and other performance analysis tools have improved substantially in recent years and they are now able to
provide the user with very accurate and descriptive interpretations of the various bottlenecks in a parallel
application.
However, most users in the scientific computing community do not have the time or expertise to study and 
tune the parameters of the communication libraries used by their
codes. In fact, optimizing the parameters of communication libraries requires technical knowledge and time to try different configurations.
For example, most Message Passing Interface (MPI) implementations offer hundreds of parameters that can provide
significant speedup if they are set to their optimal value (which varies depending on the application), compared to the default configuration.

Furthermore, general-purpose communication libraries, like MPI, express several parallel programming models
(e.g. one-sided, message-passing, task-based, etc...), and the optimal setting of a parameter used for a programming model
might impact the performance when used on a different application, using a different programming model.

On the other hand, run-time communication libraries usually express fewer parallel programming models than
general-purpose parallel programming libraries, and thus the communication pattern exposed by a run-time library
can be interpreted and modeled much more easily.

In this work, we explore the use of machine learning techniques to optimize a particular run-time communication library, 
namely the OpenCoarrays run-time (used by the GNU Fortran compiler to implement the coarray
support) and particularly its implementation on top of MPI-3.
Finding the perfect learning algorithm for AITuning is beyond the scope of this paper and
we plan to explore more the machine learning aspects of this problem in a future work, however the results
we show in Sect.\ref{sec:results} are already very good.

Another important goal of this work is to demonstrate how the MPI Tool Information Interface, introduced by the MPI-3 standard,
can be used effectively for automatic performance improvements when used by run-time
libraries based on MPI-3, such as OpenCoarrays.

\section{Related Work}

The problem of tuning and auto-tuning communication libraries, like MPI, has been tackled several times in the past,
using many different approaches.

In~\cite{AutoTune}, Miceli et al. propose AutoTune, an extension of Periscope~\cite{Periscope}, an automatic
distributed performance analysis tool. This framework tries to optimize a parallel application
under many aspects including MPI tuning, thread affinity, and CPU frequency.

In~\cite{MPI_PTF}, Sikora et al. extend again Periscope as part of the AutoTune project to implement
autotuning capabilities for MPI applications.
The output of the framework proposed is a set of tuning recommendation that can be integrated into the
production version of the code. This tool provides the user with evolutionary algorithms able to
heuristically guide the search of the most significant tuning parameters in MPI by executing a
reasonable number of experiments.

Pellegrini et al. in~\cite{AI_MPI_Tuning}
propose the use of two machine learning algorithms (decision trees and neural networks), to implement
a predictive model that analyzes any MPI input program, and according to gained knowledge of the architecture,
produces the value of a set of a predefined runtime parameters that provide optimal speedup.
The overall approach proposed by Pellegrini et al. is similar to what we describe in this work,
but our machine learning approach and modelization is completely different because it makes use of deep
reinforcement learning techniques.

\section{(Deep) Reinforcement Learning}\label{sec:DRL}


The idea behind Reinforcement Learning is to have a learner called \textit{agent} which interacts
with an \textit{environment} through \textit{actions}.
The \textit{environment} responds to the \textit{actions} and it presents new situations to the \textit{agent}.
The \textit{environment} also gives rise to \textit{rewards}: a numerical representation that the \textit{agent}
tries to maximize.
The final goal of Reinforcement Learning is to find a \textit{policy}, that maximizes the overall reward for the agent.
A \textit{policy} is a mapping from states to probabilities of selecting a certain \textit{action}.
Reinforcement Learning methods specify how the \textit{agent} changes its \textit{policy} as a result of experience.

A very important assumption made by RL systems is that the \textit{environment} and its states posses
the \textit{Markov property}; meaning that each state is expected to summarize all the past and relevant
information. If a state has the \textit{Markov property}, then the \textit{environment} response at $t+1$ depends
only on the state and action at time $t$. A RL task is also called \textit{Markov Decision Process} or MDP.

If all the elements of the MDP (probability transitions, rewards, states, actions) representing the \textit{environment}
are known, then the RL task is called \textit{model-based};
this is rarely the case in the real world, but there are very efficient ways to solve this RL task and find the optimal policy.
If there is no (or partial) knowledge of the environment, then the RL task is called \textit{model-free} and only
experience is used to find optimal policies.

The basic idea behind several model-free RL algorithms is to estimate the action-value function (Q), which expresses how
rewarding is to make an action in a particular state, by using the Bellman equation \cite{Bellman} as an iterative update.
In equation~\ref{eq:Q} we report the \textit{Bellman optimality equation} for $Q$,
where $\gamma$ represents a discount factor which indicates how much influence the future value of Q has on the current Q.

\begin{equation}
  \label{eq:Q}
  Q^*(s,a) = \mathbb{E} \{r_{t+1} + \gamma \max_{a'} Q^*(s_{t+1}, a') | s = s_t, a = a_t \}
\end{equation}



Although effective, using Equation~\ref{eq:Q} as an iterative update is impractical.
Even if we have a complete and accurate model of the environment's dynamics, it is usually not possible to compute an optimal policy by solving the Bellman optimality equation.
A much better approach is to use a function estimator for $Q$, which produces better values, thanks to the experience accumulated, even for states that have not been visited yet.
In the well known TD-Gammon paper~\cite{TD-Gammon}, a neural network is used to learn $Q$ and even though the algorithm makes bad decisions for rarely visited configurations, it makes optimal decisions for frequently visited states.
The on-line nature of reinforcement learning tasks makes it possible to approximate optimal policies in ways that
tolerate to make bad decisions in states that are rarely encounter but very good decision in states that are
frequently encountered.

\subsection{Deep Q-Learning}\label{subsec:DQN}

Q-Learning is a reinforcement learning technique. It belongs to the class of model-free methods and tries to estimate
the Q-value function using the update equation expressed in~\ref{eq:Q-learning}.

\begin{equation}
  \label{eq:Q-learning}
  Q(s_t,a_t) = Q(s_t,a_t) + \alpha \lbrack r_{t+1} + \gamma \max_{a} Q(s_{t+1}, a) - Q(s_{t}, a_t)\rbrack
\end{equation}

Q-learning is just the Bellman optimality equation applied iteratively to evaluate and improve the Q-value function
in a model-free problem, using a greedy policy. In other words, the best update rule to estimate the optimal action-value function Q for a given state,
is the quantity that leads to the optimal policy.
The optimal policy is the one given by the Bellman optimality equation, which is the max Q among all possible actions in the next state.

The Q-learning algorithm can be implemented by just keeping track of the Q-values of all the visited states in a table,
but this is prohibitive for real problem with a large number of states.

Alternatively, one could estimate the Q-value of the states, using various techniques.
One of these is called ``Deep Q-Learning'' and it involves the use of a deep neural network for the estimate.
Unfortunately, applying non-linear function approximators to model-free algorithms, such as Q-learning,
could cause the Q-network to diverge~\cite{TD-nonlinear}, however there have been works to fix the divergence issue 
such as the gradient temporal-difference methods like~\cite{Convergent_TD}
and~\cite{control_function_approx}.

The most famous and meaningful example of successful application of deep reinforcement learning is probably~\cite{AtariDRL},
where a convolutional neural network has been used to interpret the state of an Atari video game to
produce the values of Q for all the possible actions allowed by the game.
In the Atari work~\cite{AtariDRL}, the stability of the Q-learning algorithm, while using neural networks, is guaranteed by
two mechanisms: experience replay and fixed Q-targets.
Experience replay is random sampling over the entire experience accumulated and applying an optimization step on the
neural network using the samples. This mechanism makes sure to break the temporal correlation of the experience
observed by the network, resulting in a better stability and convergence of the algorithm.
Q-targets means that the Q values used to compute the updates of the Q-learning algorithm belong to a neural network
trained on old values. In~\cite{AtariDRL}, the authors use two neural networks, an they switch between the two after a
certain number of steps to compute the Q-value for the targets in the Q-learning algorithm.


\section{Potential in Communication Library Introspection}\label{sec:introspection}

Understanding the performance issues of an MPI code is an operation that requires low-level information;
for example, knowing how much time is spent in an \texttt{MPI\_Recv} can help to understand whether the application suffers of poor load balancing or just high communication costs.
Such a low-level information is usually hidden into the internal variables of the MPI implementation.
For example, a typical information that can be useful to know is \textit{how many messages are in the Unexpected Message Queue waiting to be received?}.

With the new tools information interface introduced in MPI-3, MPI provides a standard way to access performance data contained inside the MPI implementation (called \textit{performance variables}) and internal variables that control
the behavior of the implementation (called \textit{control variables}). An example of a \textit{control variable} is the one that
defines the threshold, associated with the message size, that decides whether a message should be sent using the eager or
rendezvous protocol.

Although the performance variables are common to any MPI implementation (e.g., Unexpected Message Queue length), the MPI Forum does not specify a direct way to get the status of these variables.
The intent of the MPI Tool Information Interface (from now on MPI\_T, see Section~\ref{subsec:MPI_T})
is to enable an MPI implementation to expose implementation-specific details; for this reason is not possible to define variables that all MPI implementations must provide. This approach is called \textit{introspection}.
The most common use case for the MPI\_T is to provide performance information
and control variables to profilers and debuggers in order to help the users understanding issues and bottlenecks
in MPI applications.

It is possible to write applications that take advantage of the information provided
by MPI\_T, 
but introducing such low-level concepts in user code is not advisable.
We believe that the best opportunities to improve the performance of an MPI
application using MPI\_T are in the run-time communication libraries built on top of MPI.
In fact, MPI\_T has been already successfully used by run-time communication
libraries to select the best algorithm based on the support provided by the MPI implementation.
For example, Fanfarillo and Hammond in~\cite{EventsMPI3} use the MPI\_T to
select the best algorithm to implement \textit{events} in OpenCoarrays~\cite{OpenCoarrays},
with a remarkable performance enhancement.

\subsection{MPI Tool Information Interface (MPI\_T)}\label{subsec:MPI_T}


MPI\_T provides a standard interface to access \textit{performance variables} and
\textit{control variables}.
For both types of variables, there are several common concepts. In order to access a variable, an handle must
be created first. With the handle the MPI implementation can provide low-overhead access to the internal
variable.

\textit{Control variables} allow the use to influence how the MPI implementation works.
In order to use a \textit{control variable}, the variable needs to be discovered.
MPI provides functions 
to implement \textit{introspection}, discover
how many control variables are available, getting their details and modifying  their values.
During this work, we found out that it is important to modify all the \textit{control variables} values before calling
\texttt{MPI\_Init}.

\textit{Performance variables} are usually expressed in terms of queue lengths, waiting times, re-transmission attempts.
For example, in a load imbalanced situations, where some processes make send requests before that the corresponding receives
have been posted, the length of the unexpected message queue will be longer on some processes than on others.
Another typical symptom of load imbalance is the longer time spent in a receive, waiting for the data to arrive.
By combining the data with an understanding of how the implementation works, profilers are able to provide clues to the
programmer on how to determine the source of the performance problem.
The way \textit{performance variables} are accessed is similar to the way \textit{control variables} are managed but
\textit{performance variables} require an additional step: the creation of a session.
A session enables different parts of the code to access and modify a \textit{performance variable} in a way
that is specific to that part of the code. In other words, a session provides a way to isolate the use of a
\textit{performance variable} to a specific part of the code.
In order to read the value associated with a \textit{performance variable} the creation of handle and session should be
performed after calling \texttt{MPI\_Init}.

\subsection{OpenCoarrays}\label{subsec:opencoarrays}

OpenCoarrays~\cite{OpenCoarrays} is an open-source software project for developing, porting and tuning transport layers that support coarray Fortran compilers.  
It targets compilers that conform to the coarray parallel programming feature set specified in the Fortran 2008 standard.
It also supports several features defined in the Fortran 2018 standard including: \textit{events} for fine-grain
synchronization between parallel entities, \textit{failed images} to manage failures, collective/reduction (called \textit{collective}), and a partial implementation of \textit{teams}, used to create independent subgroups of parallel entities.
Currently, it is used as the run-time communication library by the GNU Fortran (GFortran) compiler.

OpenCoarrays defines an application binary interface (ABI) that translates high-level communication and synchronization requests into low-level calls to a user-specified communication run-time library.
This design decision liberates compiler teams from hardwiring communication-library choice into their compilers and it frees Fortran programmers to express parallel algorithms once, and reuse identical CAF source with whichever communication library is most efficient for a given hardware platform.

Since the first release of OpenCoarrays (August 2014), the widest coverage of coarray features was provided by a MPI based run-time library (LIBCAF\_MPI).
Because of the one-sided nature of coarrays, the run-time library uses almost exclusively MPI one-sided communication routines based on passive synchronization.






\section{AITuning Design}\label{sec:ai-design}

AITuning has been designed as a separate component from run-time communication libraries. Its purpose
is to guide the automatic tuning process of the libraries utilizing machine learning techniques.
It is written in C++ and it is structured to be completely agnostic of run-time libraries,
communication libraries, and machine learning algorithms and paradigms (although Reinforcement Learning
approaches are well suited for this problem).


\subsection{Architecture}\label{subsec:AITuning_architecture}


\footnote{A class diagram of the architecture is available on https://github.com/NCAR/AITuning}The Controller class exposes a set of methods identified by the prefix \texttt{AITuning\_*} that
can be called by the run-time library.
The method \texttt{AITuning\_start(string layer)} 
takes a string representing the communication layer to be used. This method needs to be called
before the initialization of the communication library (in this case \texttt{MPI\_Init\_thread}).
In order to plug AITuning in OpenCoarrays without changing the source code of the latter, we decided to use
the MPI Profiling Interface.
We created wrappers for the MPI functions that AITuning needs to interact with (e.g. \texttt{MPI\_Init}
and \texttt{MPI\_Finalize}) and called the \texttt{AITuning\_*} methods from there.

In Listing~\ref{AITuning_init} we show a portion of the actual code of the \texttt{MPI\_Init\_thread}
wrapper. As explained in Section~\ref{subsec:MPI_T}, \textit{control variables} and \textit{performance variables}
needed to be set before and after the actual call to \texttt{MPI\_Init\_thread}, respectively.
Once the layer has been passed to the Controller object, a specific CollectionCreator is instantiated
using the CollectionCreator object. The actual collection (in our case MPICHCollectionCreator) has predefined lists
of control and performance variables that we decided and used for a specific AI component.

\begin{lstlisting}[caption={AITuning initialization}, captionpos=b, label=AITuning_init]
  int MPI_Init_thread(int *argc, char ***argv, int required, int *provided)
  {
    int err = -1;
    AITuning_start("MPICH");
    AITuning_setControlVariables();
    err = PMPI_Init_thread(argc, argv, required, provided);
    AITuning_setPerformanceVariables();
  }
    
\end{lstlisting}

In order to make AITuning general enough to handle any kind of control and performance variables, we decided
to declare the classes \texttt{ControlVariable} and \texttt{PerformanceVariable} as abstract.
In fact, besides the default control and performance variables defined in a specific Collection object (related
to a specific communication library implementation), it is possible to define UserDefined Performance Variables.
This class of variables allows the user to defined specific performance variables, like the time spent to
run the entire application, the time spent to execute a \texttt{MPI\_Win\_flush} and similar.
Since they all inherit from the abstract class PerformanceVariable, they can be stored in the CollectionPerformanceVar
object.
In order to read performance variables, specific objects of the class Probes should be used.
This class makes sure that the performance variables read using MPI\_T or any other way (user defined included),
respect certain criteria, like datatype, precision, and range.
In listing~\ref{AITuning_UD} we show how a UserDefined Performance Variable gets instantiated, added to the
Performance Variable collection containing the predefined \textit{performance variables}, and finally it gets
associated to a probe.

\begin{lstlisting}[caption={Declaration of UserDefined Performance Variable and Probe}, captionpos=b, label=AITuning_UD]

  UserDefinedPerformanceVar *flush_time_v = new UserDefinedPerformanceVar((char*)"flush_time",
                                           (char*)"flush_time_log.txt", 0.001);
  AITuning_addUserDefinedPerformanceVar(flush_time_v);
  flush_time_p = new SingleProbe((char*)"flush_time_probe", flush_time_v);

\end{lstlisting}

In listing~\ref{AITuning_flush} we show how to use a probe to register a performance value
(\texttt{flush\_time\_p}) and read all the performance variables listed in a Collection (including the user defined).
  
\begin{lstlisting}[caption={Performance Variable read in MPI\_Win\_Flush}, captionpos=b, label=AITuning_flush]
int MPI_Win_flush(int rank, MPI_Win win)
{
  int ret;
  double start_time_flush, end_time_flush;
  start_time_flush = MPI_Wtime();
  ret = PMPI_Win_flush(rank, win);
  end_time_flush = MPI_Wtime();

  flush_time_p->registerValue(end_time_flush - start_time_flush);
  AITuning_readPerformanceVariables();
  
  return ret;
}   
\end{lstlisting}

All the performance variables keep track of the values detected during the program execution.
At the end of the execution, in a wrapper of \texttt{MPI\_Finalize}, statistics of the values get collected
(e.g. average, max, min, median) and they will form the ``state'' representation passed to the AI component.

The entire machine learning process is performed in the \texttt{MPI\_Finalize} wrapper, at the end of the program.
The AI components receives a representation of the state of the application, which represents the state of the environment
in a reinforcement learning setting.
The reward gets computed in the AI component, based on previous data (in particular total\_execution\_time) and the reinforcement
learning algorithm gets trained on the new data and produces a new action, defined as a ``change'' for a control variable.
The new values for the control variables will be used during the next execution of the same application.
A detailed description of the training process and AI component is provided in Section~\ref{subsec:AITuning_training}.

Not all the performance variables are the same; a variable like total\_time cannot be passed to the RL algorithm
as an absolute value. In fact, the same application has very different execution times when run on a different numbers
of processes.
In AITuning it is possible to declare a performance variable as ``Relative''. During the first run, the performance
variable declared as relative will maintain in memory the absolute value of the quantity they represent.
During the other runs, all the values of a relative performance variable are express as the difference between
the absolute value obtained during the first run and the current absolute value.
For example, if we consider the total execution time as performance variables,
a positive value can be seen as a performance improvement, since during the first run the execution time was higher that the new value.
This representation allowed us to write easy reward functions based on the results of relative variables.

\subsection{Training}\label{subsec:AITuning_training}

As first step, 
all the values of the performance variables are ``standardized'' against
a reference run.
To do so, a first run (or set of runs) is used as a reference for performance variables related to
time and to a specific run in a consistent way.
For this reason, when AITuning is active, the first run of the application is used to record the performance variables of the application
when using a vanilla MPI implementation.
The user communicates the first run by setting an environment variable \texttt{AITUNING\_FIRST\_RUN = 1}.

For every run other than the first, the algorithm produces a new action in the form of a ``change'' on a control variable.
Each control variable has a fixed ``step'' to be used to change the absolute value of the control variable.
For example, the MPICH control variable \texttt{MPIR\_CVAR\_ASYNC\_PROGRESS} which controls the use of a helper thread to implement
MPI asynchronous progress, can assume only two values: 0 and 1. 
On the other hand, the variable \texttt{MPIR\_CVAR\_CH3\_EAGER\_MAX\_MSG\_SIZE} assumes a numerical value representing the message size threshold 
to switch from the eager to the rendezvous protocol: in this case AITuning will change its value in predefined steps of 1024.

In every run, the neural network in charge of estimating the Q-value produces an estimate of the Q-value given a certain state provided by the
performance variables. At the end of the run, the new reward gets computed and the neural network gets retrained based on the outcome.
In order to make the Q-learning stable, we used the replay technique described in Section~\ref{subsec:DQN}. We pick a random subset of the whole
experience accumulated every 200 runs, and we train the neural network on that. We have not implemented the Q-target technique.

\subsection{Control and Performance Variables for MPICH}\label{subsec:ControlPerfVar}

For now, we focused our efforts only MPICH-3.2.1 because of the small
number of control and performance variables exposed by the implementation,
which made our reinforcement learning algorithm design and training faster.
The control variables chosen for MPICH-3.2.1 are
\verb|ASYNC_PROGRESS|, \verb|CH3_ENABLE_HCOLL|, 
\verb|CH3_RMA_DELAY_ISSUING_FOR_PIGGYBACKING|, \\ 
\verb|CH3_RMA_OP_PIGGYBACK_LOCK_DATA_SIZE|, 
\verb|POLLS_BEFORE_YIELD|, \verb|CH3_EAGER_MAX_MSG_SIZE|.
The only performance variable chosen from MPICH-3.2.1 was \texttt{unexpected\_recvq\_length}, representing the length of the
unexpected message queue. We use several user-defined performance variables related to the average and maximum time needed to complete
\texttt{MPI\_Win\_Flush}, \texttt{MPI\_Put},\texttt{MPI\_Get}, and total application time.
We also added the number of processes used in the run as input parameter.

\subsection{Inference}\label{subsec:AITuning_inference}

AITuning will be shipped along with OpenCoarrays already trained for several MPI implementation and transport layers (e.g. GASNet).
When the user decides to activate AITuning, he/she will compile OpenCoarrays using the PMPI wrapper.
At this point, we recommend the user to run their application for at least 20 times. During these 20 runs, the RL algorithm will 
``explore'' the new application
and produce the right combination of parameters. During this exploration phase, AITuning may produce a configuration that penalizes the performance.
At the end of the 20 runs, AITuning analyzes the results, discards the runs where the performance was penalized, and applies the median over the
values of the control variables of the runs that provided good results within 5\% from the best (creating an ensemble).

\subsection{Convergence of the Reinforcement Learning}\label{subsec:AITUning_RL}

We ran a number of simulations to assess the performance of our choices for the implementation of Reinforcement Learning, to assess
if it were able to find (converge to) an optimal value for all control variables. In these simulations,
there was no OpenCoarray library to tune, just models. Each model included a handful of simulated control and performance
variables with known behavior and added Gaussian noise (to simulate run-to-run variability). An example of a simulated performance variable
we used is a function of one control variable, for example in the shape of a parabola, with a global minimum.
Even with high level of noise (up to 30\% of the value of the performance variables), our algorithm
has always been able to find a set of control variables reasonably close to the known best. In a future work, we intend to
explore this aspect in greater details, utilizing more complex simulated performance variables (for example depending on more than
one control variable).


\section{Experimental Evaluation}\label{sec:exp-eval}

In order to train AITuning properly on MPICH-3.2.1, we decided to use two different
supercomputers: Cheyenne (NCAR) an SGI machine with InfiniBand network interconnect
and Edison (NERSC) a Cray XC30 with Aries interconnect.
For the training we decided to use four main codes parallelized with Coarrays Fortran:
1) CloverLeaf~\cite{CloverLeaf}, 2) Lattice-Boltzmann code~\cite{lbm_rosales},
3) Skeleton Particle-in-cell~\cite{PIC-CAF}, 4) Parallel Research Kernels~\cite{PRK}.
We have run the aforementioned codes using a different number of processes going from 64 to 2048
for a total of 5000 runs.

\subsection{Intermediate Complexity Atmospheric Research}\label{subsec:ICAR}

The Intermediate Complexity Atmospheric Research (ICAR)~\cite{ICAR} model developed at NCAR,
is a simplified atmospheric model designed primarily for climate downscaling,
atmospheric sensitivity tests, and educational uses.
ICAR is a quasi-dynamical downscaling approach that uses simplified wind dynamics to perform high-resolution meteorological
simulations 100 to 1000 times faster than a traditional atmospheric model and can therefore be used to better characterize uncertainty across numerical
weather prediction models and climate models, and in dynamical downscaling.

In~\cite{CAF-ICAR}, Rouson et al. developed a mini-app of the ICAR model using coarray Fortran, showing great performance improvements.
Since then, lead developer of ICAR, Ethan Gutmann, developed a fully functional version of ICAR based on coarray Fortran, which we used
for testing AITuning.
The version of ICAR we used is a full atmospheric model; the code include computation, communication and IO parts.

\subsection{Results Evaluation}\label{sec:results}

\begin{figure}
 \includegraphics[scale=0.6]{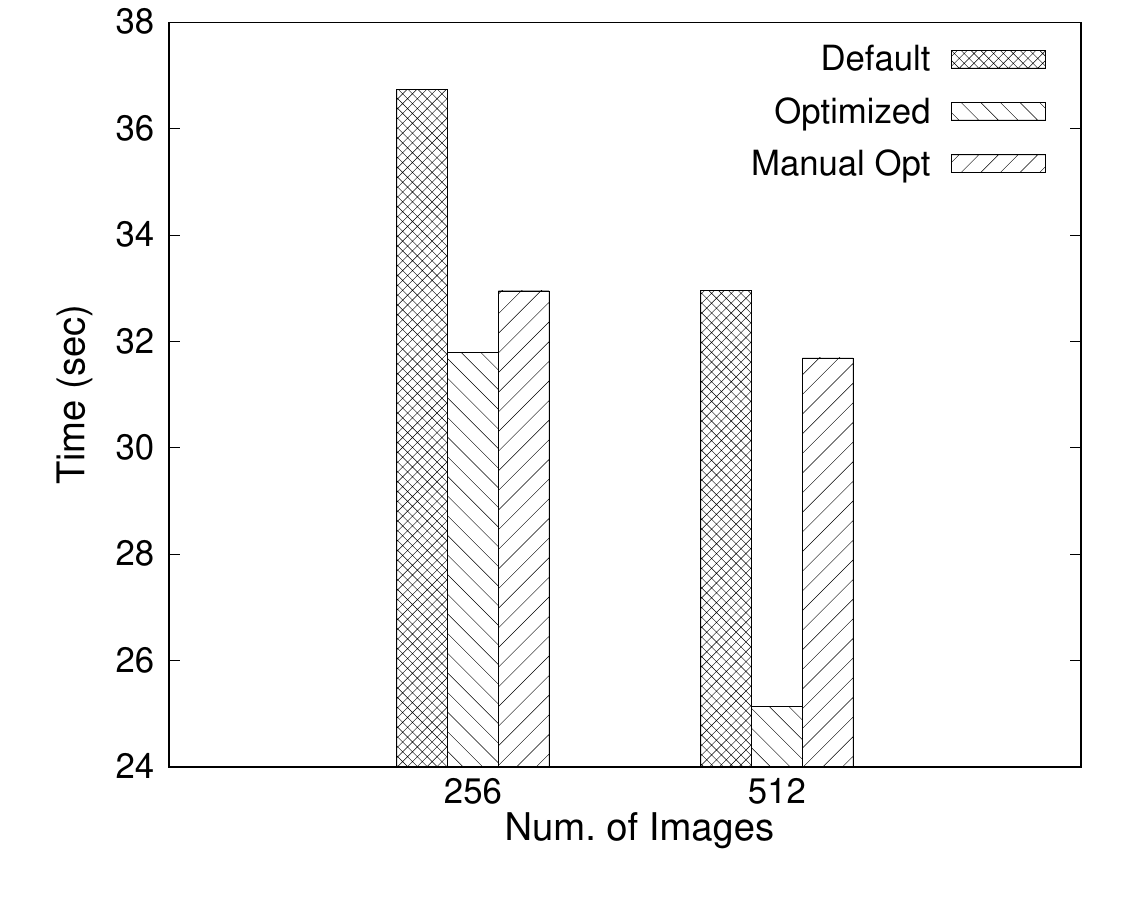}
  \caption{Performance comparison between default and optimized configurations}
  \label{fig:icar-results}
\end{figure}

In Figure~\ref{fig:icar-results}, we report the results obtained for ICAR running on Cheyenne using the default ``vanilla'' configuration set in MPICH-3.2.1,
the optimized configuration found by AITuning after running ICAR 20 times, and an human optimized version based on reasonable guesses.
The ``default'' bars represent the total time needed to complete a test case on ICAR using the default settings and in both cases, with 256 and 512 images,
it provides the worst performance. On the other hand, the ``optimized'' version produced by AITuning always leads to the best performance.
In both the 256 and 512 images cases,
the manual optimization increased the eager limit by an order of magnitude higher than the default while leaving all the other setting as in the default
configuration.
For the case with 256 images, the optimized version provides 13\% performance improvement compared to the vanilla version.
For the case with 512 images, the optimized version provide 25\% performance improvement over the vanilla version, mostly because of the
higher communication cost imposed by the higher number of processes and same problem size (strong scaling).

The most influential tuning parameter for the ICAR test case resulted to be the presence of the asynchronous progress thread.
We also noticed that some parameters have a different influence based on the number of processes being used.
In particular, the value of \texttt{MPICH\_POLLS\_BEFORE\_YIELD} played a much more relevant role in the case with 512 images than in the case with 256 images.
This is not surprising because ICAR attempt to overlap computation with communication by using coarray ``puts'' instead of ``gets''.
For the 256 case, the optimal configuration found by AITuning had \texttt{MPICH\_POLLS\_BEFORE\_YIELD} set to the default value 1000, meaning that
it was found not relevant. On the other hand, for the 512 images case, AITuning found a value of 1100.
We manually changed the value of \texttt{MPICH\_POLLS\_BEFORE\_YIELD} by keeping the configurations found by AITuning the same for both cases and found that
in the case with 512 images, a value of \texttt{MPICH\_POLLS\_BEFORE\_YIELD} between 1200 and 1500 provides the best performance, so it seems
there is still room for improvement.

\section{Conclusions and Future Work}

In this work, we presented AITuning, a machine learning-base tuning tool for run-time libraries.
AITuning has been released under open-source license and it is currently available on github \footnote{https://github.com/NCAR/AITuning}.
It currently works with the OpenCoarrays library, but its structure allows it to be extended to any run-time communication
library, based on any communication layer.
To the best of our knowledge, this paper is a unique contribution because it is the first attempt to try 
to find the optimal tuning parameters used a deep reinforcement learning algorithm and MPI\_T.
We tested AITuning and our reinforcement learning algorithm, carefully designed for MPICH-3.2.1, using a real atmospheric code: ICAR.
AITuning was able to produce a configuration of parameters that lead to 13\% and 25\% performance improvement for the case running
on 256 and 512 images, respectively.

In the future, we plan to extend our analysis to other MPI implementations with a higher number of control and performance variables.
Furthermore, we will explore more options on the reinforcement learning algorithm, and potentially other machine learning approaches.





\bibliographystyle{ACM-Reference-Format}


\end{document}